\documentclass[letterpaper, 10 pt, conference]{ieeeconf}  % Comment this line out if you need a4paper

\IEEEoverridecommandlockouts                              % This command is only needed if 
\usepackage[pagebackref=true,breaklinks=true,colorlinks,citecolor=green]{hyperref}
\usepackage[utf8]{inputenc} % allow utf-8 input
\usepackage[T1]{fontenc}    % use 8-bit T1 fonts
\usepackage{hyperref} 
\usepackage{graphicx}
\usepackage{url}            % simple URL typesetting
\usepackage{booktabs}       % professional-quality tables
\usepackage{amsfonts}       % blackboard math symbols
\usepackage{nicefrac}       % compact symbols for 1/2, etc.
\usepackage{microtype}      % microtypography
\usepackage{paralist}
\usepackage{xcolor}
\usepackage[utf8]{inputenc} % allow utf-8 input
\usepackage[T1]{fontenc}    % use 8-bit T1 fonts
\usepackage{hyperref}       % hyperlinks
\usepackage{url}            % simple URL typesetting
\usepackage{booktabs}       % professional-quality tables
\usepackage{amsfonts}       % blackboard math symbols
\usepackage{nicefrac}       % compact symbols for 1/2, etc.
\usepackage{microtype}      % microtypography
\usepackage{paralist}
\usepackage{amsmath}
\let\labelindent\relax
\usepackage{enumitem}

\title{OPEn: An Open-ended Physics Environment for Learning Without a Task}

\author{Chuang Gan$^{1,2}$, Abhishek Bhandwaldar$^{2}$,  Antonio Torralba$^{1}$,  Joshua B. Tenenbaum$^{1}$, Phillip Isola$^{1}$  \thanks{
 $^1$ Massachusetts Institute of Technology,
 $^2$ MIT-IBM Watson AI Lab, Project page: \protect\url{http://open.csail.mit.edu/}}}
\begin{document}

\maketitle

\begin{abstract}

Humans have mental models that allow them to plan, experiment, and reason in the physical world. How should an intelligent agent go about learning such models? In this paper, we will study if models of the world learned in an open-ended physics environment, without any specific tasks, can be reused for downstream physics reasoning tasks. To this end, we build a benchmark {\bf O}pen-ended {\bf P}hysics {\bf En}vironment (OPEn) and also design several tasks to test learning representations in this environment explicitly. This setting reflects the conditions in which real agents (\emph{i.e.} rolling robots) find themselves, where they may be placed in a new kind of environment and must adapt without any teacher to tell them how this environment works. This setting is challenging because it requires solving an exploration problem in addition to a model building and representation learning problem. We test several existing RL-based exploration methods on this benchmark and find that an agent using unsupervised contrastive learning for representation learning, and impact-driven learning for exploration, achieved the best results. However, all models still fall short in sample efficiency when transferring to the downstream tasks. We expect that OPEn will encourage the development of novel rolling robot agents that can build reusable mental models of the world that facilitate many tasks. 

%Mental models allow humans to plan, experiment, and reason in the physical world. How should an intelligent agent go about learning such a model? In this paper, we will study if models of the world that are learned in an open-ended physics environment, without any specific tasks, could be reused for downstream tasks. To this end, we build a benchmark Open-Ended Physics Environment ( OPEn) and also design several downstream tasks to explicitly test learning models and representations in this environment. This setting more accurately reflects the conditions in which real agents -- animals and robots -- find themselves, where they may be placed in a new kind of environment and must adapt without any teacher to tell them how this environment works. But it is also more difficult because it requires solving an exploration problem in addition to a model building problem. We test several existing model-based explorations agents on this benchmark and find that an agent using unsupervised contrastive representation learning for model building and impact-driven exploration achieved the best results. But all models still fall short in sample efficiency when transferring their model to the downstream tasks. We expect that  OPEn will encourage the development of novel agents that could build  reusable mental models of the world across many tasks downstream tasks.

\end{abstract}

\section{Introduction}

Reinforcement learning (RL) excels at specialist intelligence. When given a narrowly defined task, and a well-shaped reward function, training standard RL algorithms with a large amount of task-specific environment interactions can achieve impressive performance, yielding agents that beat humans at many video games~\cite{mnih2015human} and board games~\cite{silver2016mastering}. Typically, however, the agent that masters one game, say Go, might have no idea how to make sense of a different game, such as Space Invaders. In contrast, human intelligence works differently.  From infancy, humans have a mental model to infer the physical structure of the world, which is essential for making physical inferences about the world, predicting what will happen next, and planning actions. 

Motivated by the age-old idea that model-building is critical to general intelligence, jointly learning a model and exploring the environment has received considerable recent attention~\cite{oudeyer2007intrinsic,ha2018world,kaiser2019model,sekar2020planning}. However, this challenge remains a mostly open research problem. Unlike past work that built models via either random exploration~\cite{ha2018world} or downstream task optimization~\cite{kaiser2019model}, a more fundamental question is: how should an agent go about learning a useful model of the world when it is placed in an environment without an explicit task?

Although open-ended learning is a classic problem, most current RL environments are oriented toward learning to solve a specific task. For example, the clear goal in game environments like ALE~\cite{bellemare2013arcade} is to achieve a high score in the game. Although such environments can and have been used to study task-agnostic learning (e.g., \cite{pathak2017curiosity}), doing so leaves the concern that the game task is so intimately baked into the environment that it inevitably shapes learning. More recently, research has extended toward environments that support \emph{families} of tasks~\cite{yu2020meta}, such as random variations of physical parameters in robot simulations~\cite{finn2017model} or procedurally generated levels of a game~\cite{cobbe2019leveraging,risi2019procedural}. Nonetheless, these families are still typically focused on meta-task, such as ``pick and place", ``maze navigation," or ``killing enemies in a platformer". We instead propose an environment that is designed from the ground up to be a ``sandbox", where the implicit meta-task is simply to learn about how a physical world works.

%\emph{(a difficulty here is that we have the downstream tasks, should somehow mention that)}

To this end, we build an Open-ended Physics Environment (OPEn) to combine the idea of active learning, whereby a learner gets to pick the datapoints it trains on, and world modeling, where a learner attempts to model the dynamics of an environment. The design of this benchmark platform reflect two perspectives that we believed are essential for model building:  
\setdefaultleftmargin{1em}{1em}{}{}{}{}
\begin{compactitem}
\item  \textbf{Active exploration.}  Active exploration has the chance to be more efficient than passive learning since training data can be selected based on what would be most informative to the agent at any given moment. But more than this, passive learning is simply not an option in many settings. Passive learning relies on the agent being fed a data stream to learn from. Often this data stream is a highly instructive teaching signal, as is the case in so-called ``unsupervised" learning from human-curated datasets like Imagenet~\cite{russakovsky2015imagenet}. In ecological settings, agents are not given Imagenet -- they must explore the world to seek out useful data to learn from.
\item  \textbf{Physical interaction.}  Interacting with an environment is a form of intervention that allows us to understand how their actions could affect the world and then build causal models of the world.
\end{compactitem}

\begin{figure*}[t]
   \centering
   \includegraphics[width = 0.8\linewidth]{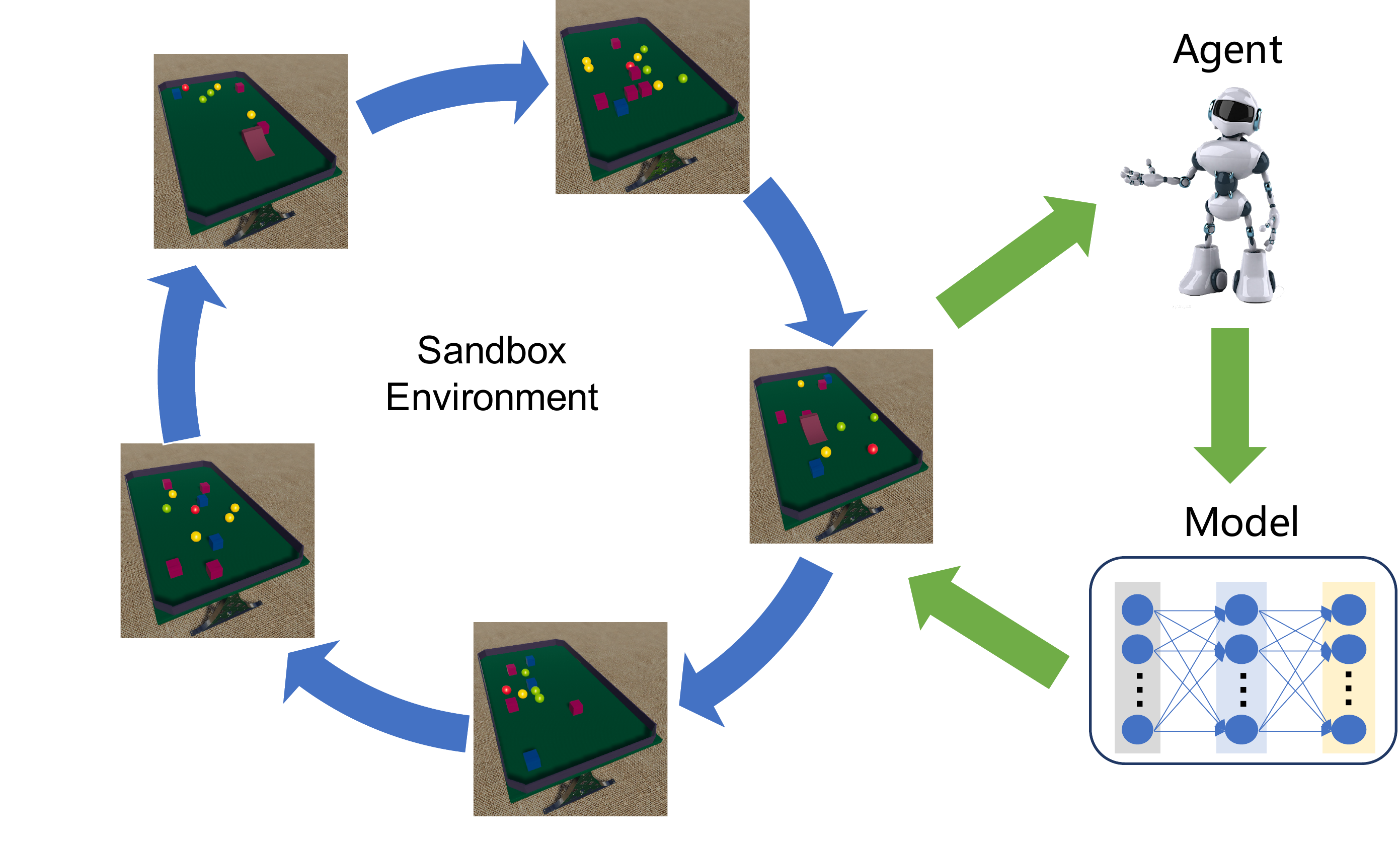}
 \vspace{-2mm}
   \caption{Problem setup: a rolling robot agent is allowed to explore the sandbox-like 3D physical world by interacting with objects without any pre-defined tasks or extrinsic rewards. The green and blue loops indicate learning within an environment and swapping to a new environment, respectively.}
   \label{fig:task}
   \vspace{-2mm}
\end{figure*}
%\footnotetext{Figure credit: \url{ https://www.pinterest.com/pin/682858362226900589/}}

As shown in Figure~\ref{fig:task}, our environment includes a sandbox for learning, and a suite of evaluation metrics to probe what was learned. In the sandbox, the rolling robot agent is allowed to explore the world by interacting with objects. There are no pre-defined tasks or extrinsic rewards. The rolling robot must use its intrinsic motivations to actively explore the environment and build a model of its. 

For evaluation, we propose an initial suite of tasks that may probe what was learned. On each task, the agent finetunes its policy using extrinsic rewards associated with that task. The intent is not that these evaluation tasks become the learning targets themselves, but rather that they just reveal what was learned in the open-ended phase. As such, this suite should be expanded over time, so that users of the benchmark do not overfit to any particular set of evaluation tasks. Moreover, methods that use our benchmark should not be solely measured in terms of how well they do on the evaluation tasks, but also on whether or not they ``cheated" by explicitly turning themselves toward the evaluation tasks. This may seem a delicate balance at first glance -- how can we really tell that an algorithm was agnostic to the evaluation suite -- but we note that this approach has lead to significant progress in unsupervised and self-supervised learning, where the goal is to develop \emph{general-purpose} representations but the standard practice is, like ours, to evaluate by finetuning on a suite of \emph{specific tasks}.

We test several existing RL-based agents on this benchmark by transferring their learned representations to the tasks in our evaluation suite and find that an agent using unsupervised contrastive representation learning for model building, and impact-driven exploration, achieved the best results. However, all models still fall short in sample efficiency when transferring their representations to the downstream tasks. To sum up, our work makes the following contributions:
\setdefaultleftmargin{1em}{1em}{}{}{}{}
\begin{compactitem}
    \item We introduce a new physical reasoning task, focusing on open-ended learning with active interactions and task-agnostic model-building.
    
    \item We build a 3D physical environment with fairly photo-realistic images and provide high-fidelity physical simulation to facilitate research on embodied intelligence in more challenging scenarios.

    \item We test several agents on this benchmark and find an agent that adopts impact-driven exploration, and contrastive unsupervised representations learning achieves the best results. All models fall short on sample efficiency. 
\end{compactitem}

\section{Related Work}

\begin{table}[t]
\small

	\begin{center}
	\begin{tabular}{lccccccc}
	\toprule
      Dataset & Realisic & 3D  &  Interactive
         & Explicit task \\
   \midrule          
Phyre~\cite{bakhtin2019phyre} & $\times$ & $\times$ &  \checkmark & \checkmark \\
             \midrule 
          IntPhy~\cite{riochet2018intphys} &  \checkmark &  \checkmark & $\times$& \checkmark \\
          \midrule    
        CATER~\cite{girdhar2019cater} & \checkmark & \checkmark & $\times$ & \checkmark\\
      \midrule    
         CoPhy~\cite{baradel2019cophy} & \checkmark & \checkmark & $\times$ & \checkmark\\
         \midrule
        CLEVRER~\cite{yi2019clevrer} & \checkmark & \checkmark  & $\times$ & \checkmark \\
         \midrule
         OPEn & \checkmark & \checkmark & \checkmark & $\times$  \\
    \bottomrule
	\end{tabular}
	\end{center}
	\caption{Comparison between OPEn and other physics benchmarks. }
\vspace{-5mm}

	\label{tab:dataset_comparison}

\end{table}

\noindent \textbf{Benchmarks for Physics Reasoning.} Our work is in the domain of physical scene understanding~\cite{battaglia2013simulation,agrawal2016learning,innamorati2019neural,huang2021plasticinelab,chen2021grounding,gan2021threedworld,ehsani2020use}.
Recently, several datasets have been developed for physics reasoning.  Intphys\cite{riochet2018intphys} proposed a synthetic dataset for visual intuitive physics reasoning. CATER~\cite{girdhar2019cater} introduced a video dataset for compositional temporal reasoning.  COPHY~\cite{baradel2019cophy} studied counterfactual physical dynamics prediction. CLEVRER~\cite{yi2019clevrer} investigated casual reasoning in collision events and also grounding it to the language domain.  Phyre~\cite{bakhtin2019phyre} designed several puzzles to examine agents' ability to use tools.  As summarized in Table~\ref{tab:dataset_comparison}, our new benchmark goes beyond these existing datasets in two ways: 1) it is interactive, thereby supporting open-ended exploration (Phyre is also interactive, but 2D), 2) the learning phase is task-agnostic with no explicit goals, rewards, or supervision provided.

\noindent \textbf{Open-ended Learning.}
Open-endedness is the problem of creating systems that develop ever-increasing abilities over time, rather than saturating when they manage to solve a narrowly defined task~\cite{lehman2008exploiting}. The evolution of life on Earth is the prototypical example of open-endedness in action. Human learning and culture also seem to exhibit open-endedness -- we are ever striving toward greater understanding and control over our environment. Open-endedness has recently become a popular topic in the machine learning community, with the striking success of self-play algorithms like AlphaGo~\cite{silver2017mastering}. Many other multiagent environments have also demonstrated some level of open-ended, emergent behaviors (e.g., \cite{sims1994evolving, suarez2019neural, baker2019emergent}). In the single-agent setting, research on open-ended learning has focused on the problem of intrinsic motivation. Rather than learning from a task or from external rewards, intrinsically motivated agents have to make up their own tasks and goals. Commonly, this is formalized as novelty search or curiosity (e.g., \cite{schmidhuber2006developmental,cully2015robots,oudeyer2007intrinsic,oudeyer2009intrinsic,haber2018learning,lehman2008exploiting,lehman2011abandoning,pathak2017curiosity,burda2018exploration}). While most work in this area has focused on exploring a fixed environment, the idea has also been applied to procedurally generating a curriculum of environments~\cite{wang2019paired, cartoni2020real,openai2019rubiks}. Nonetheless, standard benchmarks for open-ended learning are lacking, which is what we attempt to address with this paper. In addition, there is little work that evaluates whether models and skills learned through open-ended exploration are useful for downstream tasks (an exception is~\cite{cartoni2020real}). We address this aspect via our suite of evaluation tasks.

\noindent \textbf{Model-based RL}
Model-based RL has received considerable attention in recent years. Several works have shown that learning an internal model of the world can significantly improve data efficiency in RL ~\cite{oh2015action,chiappa2017recurrent,ha2018world,nagabandi2018neural,kaiser2019model}.  For example, \cite{oh2015action} uses video prediction to improve sample-efficiency on Atari games. \cite{leibfried2016deep} further predicts the reward of the environment to improve exploration. \cite{ha2018world} proposed to train a generative model as a world model of the environment, and then use the learned features as inputs to train a simple policy to solve Vizdoom and 2D racing games.  Most recently, several works have shown that contrastive learning~\cite{srinivas2020curl} and data augmentation~\cite{kostrikov2020image,laskin2020reinforcement} for representation learning can also be helpful for planning and control tasks. However, 
little work has been tested on open-ended exploration in a 3D physical world, which our work studies.

\section{Environment}
\subsection{Platform}

We build a 3D physical world on top of the TDW platform~\cite{gan2020threedworld}, which consists of a graphics engine, a physics engine, and an interaction API.

\noindent \textbf{Graphic engine.} TDW adopt Unity's game-engine technology to create a 3D virtual environment. The environment uses a combination of global illumination and high-resolution textures as well as the underlying game engine's capability to produce near-photo realistic rendering in real-time. We use directional lighting to simulate a sun-like light source and customize it by changing the angle and elevation of the source. We adopt a Unity camera object capable of giving multi-sensory data such as RGB images, segmentation masks, normal and depth maps. The camera object can be moved and rotated in 3D space.

\noindent \textbf{3D Model.} Our environment consists of a range of 3D objects, including primitives, shapes, and models with high-resolution colliders and materials. The model processing pipeline can generate object colliders using V-HACD\cite{mamou2009simple}, which gives fast and accurate collision behavior. It also allows users to import their custom models and customize their visual appearance, rigid body dynamics, object-to-object dynamics, and limited object-to-fluid dynamics. We use a 3D table model with the top surface bounded by re-scaled cube walls as the base of our scene. The object primitives in our scene consist of spheres, cubes, and ramps. 

\noindent \textbf{Physic simulation.} We utilize Nvidia's Physx physics engine for simulating rigid body dynamics. We can set different physical parameters at an object level, which directly affects the object's interaction with the environment. We use a combination of force and drag in our scene, which is applied to the object's center of mass, to move it in a controlled manner. Our platform also supports pausing and stepping of the environment by a specific number of physics steps.

\noindent \textbf{Interaction API.} 
To support interaction with the environment, we provide a python API to send and receive commands and data from the Unity environment. The API is flexible enough to give users a fine-grain control over creating a scene, ranging from defining the scene layout to customizing an object's physical and visual properties. Therefore, the users could procedurally generate scene layouts based on their specified configuration. The API also helps control the simulation by allowing a series of commands to manipulate the object and advance physics in the scene. Users can also define the type of data the environment sends and includes visual information like images and segmentation masks, state data like object location, velocity, and interaction data like collisions. We integrate this API with the OpenAI gym~\cite{brockman2016openai} to facilitate training reinforcement learning algorithms.  

\begin{figure*}[t]
   \centering
   \includegraphics[width = 0.9\linewidth]{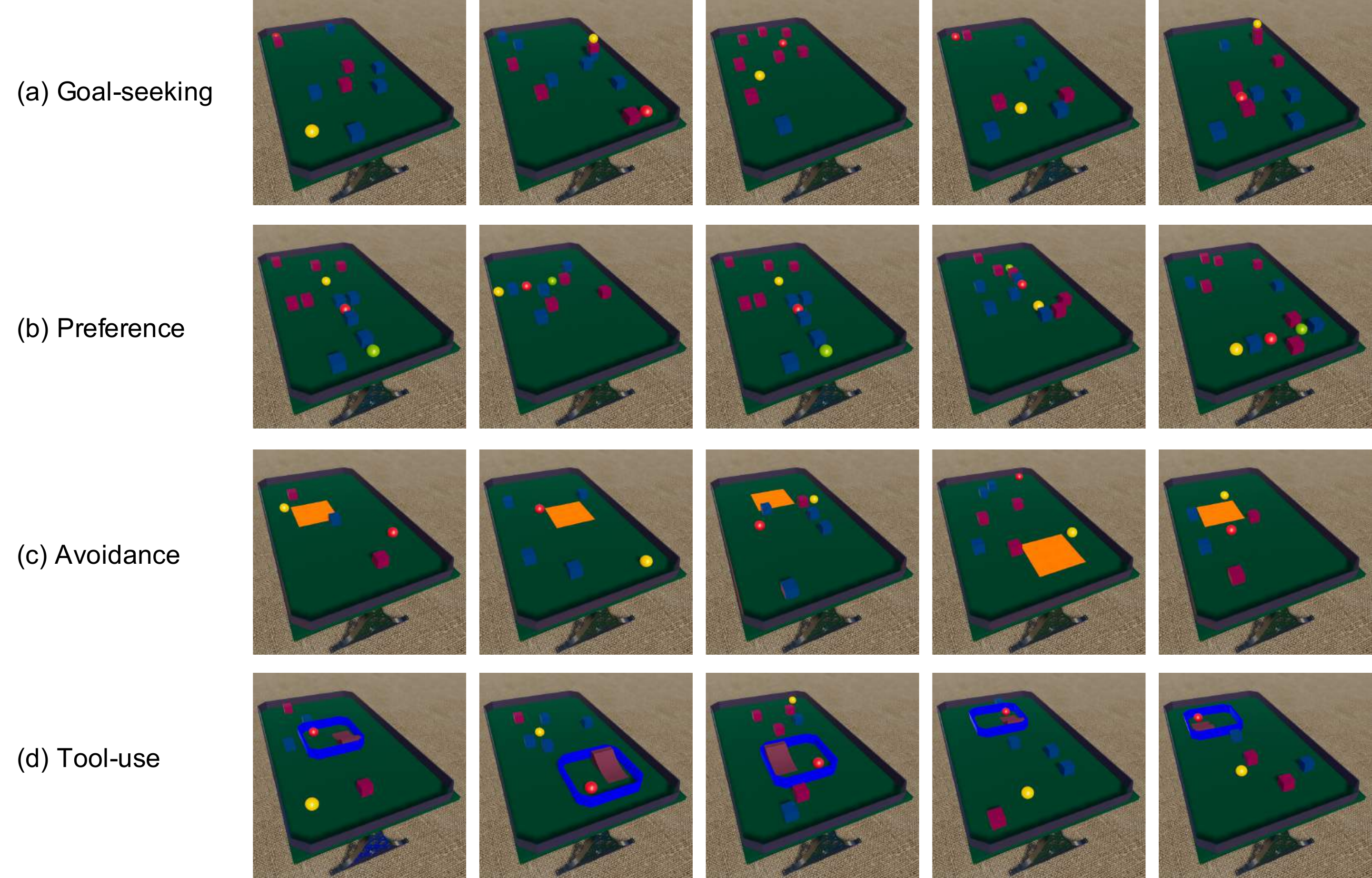}
      \vspace{-2mm}

   \caption{Evaluation suite for 4 downstream physical reasoning tasks.}
   \vspace{-2mm}
    \label{fig:puzzle}

\end{figure*}

\subsection{Problem Setup}

\noindent \textbf{Scenario.} Our scene is composed of a table with its top bounded by wall to prevent objects from falling off. The objects contained in the scene are heavy cubes (blue), light cubes (purple), three types of the sphere that act as an agent (red), goals with different rewards (yellow and green), and a ramp which can be used as a tool. The agent is capable of executing actions to move in one of the eight directions for a fixed distance and interacting with the objects in the scene by colliding with them.

\noindent \textbf{Open-ended exploration.} Our environment supports a sandbox mode that can procedurally generate puzzles with different objects placed in the scene. During open-ended exploration, we create several environments with distinctive, randomly generated configurations and select one at a time for exploration. The agent is not provided any explicit task or any extrinsic reward. The environment is non-episodic. In such an environment, the agent must learn to explore and probe different objects to learn reusable representations and models of the world. The learning algorithm is also free to switch between different environment on demand, allowing the study of learners that not only explore a single environment but also can create curricula of multiple procedurally generated environments. In practice, in the methods we compare, we switch environments whenever the loss of the world model drops below a certain threshold. When this happens we switch to the environment with the highest loss among the current pool (a form of curiosity over environment configurations). Exploration terminates when no new environment gives the agent a loss higher than a threshold.

% The action set of this environment comprises the following: \textit{Move Forward}, \textit{Move Backward}, \textit{Move Right}, \textit{Move Left}, \textit{Move Right Forward},\textit{Move Left Forward}, \textit{Move Right Backward},  \textit{Move Left Forward}, and \textit{Stop}.

\noindent \textbf{Evaluation suite.}
To evaluate what was learned by the model, we use a suite of intelligence tests inspired from~\cite{beyret2019animal}, meant to probe the degree to which the agent has acquired general-purpose representations and skills. 
%The agent needs to adapt their learned model or representation to many tasks. To evaluate the model-building abilities, we introduce four evaluation tasks.

\setdefaultleftmargin{1em}{1em}{}{}{}{}
\begin{compactitem}
\item \emph{Goal-seeking}. This category
tests the agent's basic ability to understand goals and perform action towards them. As shown in figure~\ref{fig:puzzle} (a), there is one yellow sphere and several obstacle cubes on a table. The agent is required to hit the yellow sphere as fast as possible.

\item \emph{Preferences}.  This category tests an agent's ability to choose the most rewarding course of action. As shown in figure~\ref{fig:puzzle} (b), the agent is presented with two spheres (i.e., yellow and green) and several obstacle cubes in the scene. Hitting the green sphere will earn more reward than hitting the yellow sphere. The agent is required to earn as much reward as possible with a limited interaction budget.

\item \emph{Avoidance}. This category identifies an agent's ability to detect and avoid negative stimuli. As shown in figure~\ref{fig:puzzle} (c), the goal is still to hit the yellow sphere, but there is an orange region on the table. The agent will immediately die if it enters this region. To achieve success in this task, the agent needs to understand the goal and the cost.

\item \emph{Tool-use}. This category
tests the agent's ability on a more challenging task: using tools to achieve goals. As shown in figure~\ref{fig:puzzle} (d), the agent is trapped in a fenced region and has to use the ramp to get out and hit the yellow sphere.  The agent is required to complete the task as fast as possible. 
\end{compactitem}

\begin{table*}[t]
\centering
\large
\begin{tabular}{lcccc}
Approach &  Goal-seeking & Preferences & Avoidance & Tool-use             \\ \toprule
PPO (From scratch)  &   0.509$_{\pm 0.009}$ & 0.351$_{\pm 0.02}$ & 0.185$_{\pm 0.01}$ & 0.113$_{\pm 0.08}$ \\
\midrule
ICM &   0.508$_{\pm 0.016}$ & 0.344$_{\pm 0.01}$ & 0.176$_{\pm 0.023}$ & 0.118$_{\pm 0.013}$ \\
ICM + RIDE &  0.486$_{\pm 0.005}$ & 0.367$_{\pm 0.011}$ & 0.178$_{\pm 0.019}$ & 0.120$_{\pm 0.012}$ \\
\midrule
RND &   0.470$_{\pm 0.01}$ & 0.266$_{\pm 0.02}$ & 0.168$_{\pm 0.01}$ & 0.106$_{\pm 0.01}$ \\
RND + RIDE &  0.509$_{\pm 0.014}$ & 0.372$_{\pm 0.0129}$ & 0.179$_{\pm 0.018}$ & 0.116$_{\pm 0.008}$ \\
\midrule
CURL + Random & 0.149$_{\pm 0.02}$ & 0.007$_{\pm 0.05}$ & 0.103$_{\pm 0.01}$ & 0.045$_{\pm 0.01}$\\
CURL+ RIDE & \textbf{0.532$_{\pm 0.006}$} & \textbf{0.386$_{\pm 0.04}$} & \textbf{0.191$_{\pm 0.02}$} & \textbf{0.125$_{\pm 0.02}$} \\ \midrule
Human &  0.584$_{\pm 0.008}$ & 0.588$_{\pm 0.015}$ & 0.453$_{\pm 0.025}$ & 0.363$_{\pm 0.023}$ \\
    \bottomrule
\end{tabular}
\caption{A-success scores of different method on different tasks.} 
\vspace{-5mm}
\label{tab:result}
\end{table*}
\section{Experiments}

\subsection{Compared Methods}

We implement seven agents that use intrinsic motivation for exploration and model learning, and transfer their learned representations to the tasks in our evaluation suite.% with extrinsic reward function.

\noindent \textbf{ICM.} Intrinsic Curiosity Module (ICM)~\cite{pathak2017curiosity} is a curiosity-driven exploration strategy. It adopts the prediction errors of a forward and inverse dynamics model as an intrinsic reward, which encourage the agent to take actions that improve
its ability to predict the consequence of its actions. 

\noindent \textbf{RND.} Random Network Distillation (RND)~\cite{burda2018exploration} is a counts-based exploration algorithm. An agent learns a model that predicts the feature representations of its current state extracted from a fixed randomly initialized network. The agent is encouraged to visit more unseen states and improve its coverage.

\noindent \textbf{ICM + RIDE.} Rewarding Impact-driven exploration (RIDE)~\cite{raileanu2020ride} defines a novel intrinsic reward that encourages an agent to take actions that can change the representation of the environment state. Instead of directly using the prediction error of the dynamics model as an intrinsic reward, they propose an impact-driven bonus measured by the distances between the current and next state in latent feature space. Larger state changes lead to higher rewards. We use ICM for model building and impact-driven rewards for exploration.

\noindent \textbf{RND + RIDE.} We use RND for model building and RIDE for exploration.

\noindent \textbf{CURL + RIDE.} CURL~\cite{srinivas2020curl} adopts contrastive unsupervised representations learning~\cite{he2019momentum} for model-building and shows promising results on off-policy control on top of the extracted features for many plannning and control tasks. We adapt this baseline to our problem. In particular, we use CURL for model-building and RIDE for exploration.

\noindent \textbf{CURL + Random.} We use data collected from a random policy as input to CURL.

\noindent \textbf{Plain PPO.} PPO \cite{PPO} is a SOTA policy gradient method for model-free RL. We train it from scratch.

\begin{figure*}[!h]
   \centering
   \includegraphics[width = 0.94\linewidth]{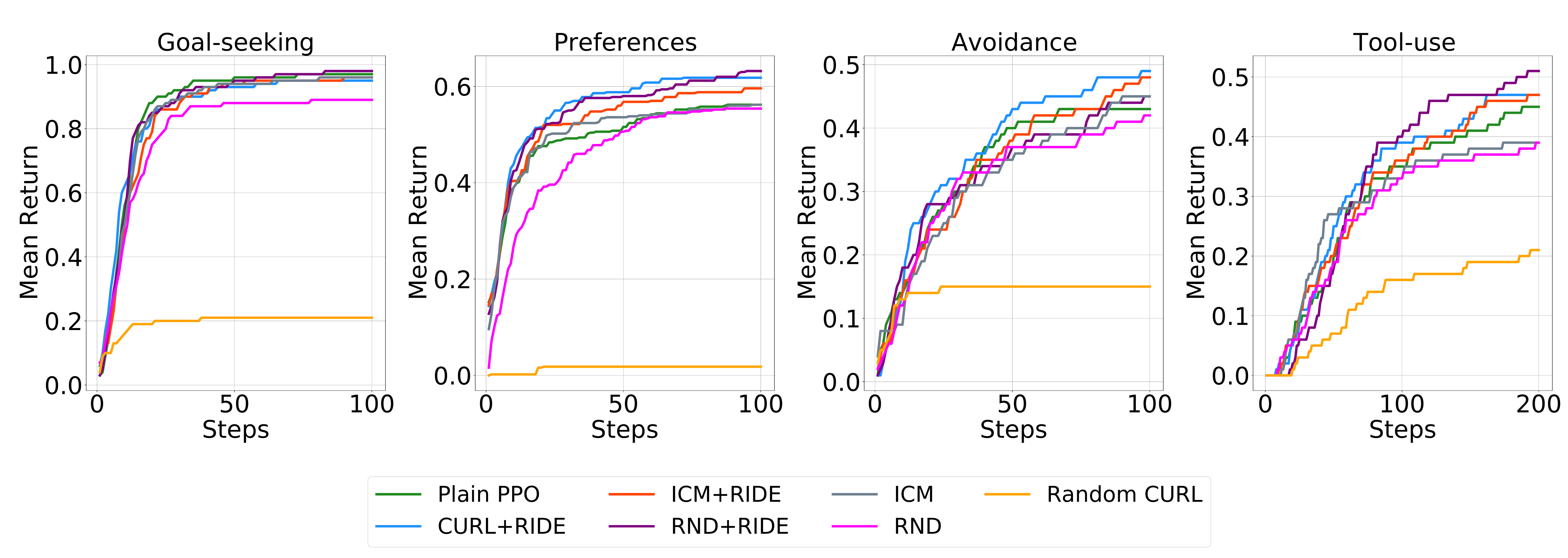}
\vspace{-2mm}
   \caption{Mean return of different models as a function of number of interactions with the environment during a single test episode.}
    \label{fig:overall}
\vspace{-2mm}
\end{figure*}
\begin{figure*}[!h]
   \centering
   \includegraphics[width = 0.94\linewidth]{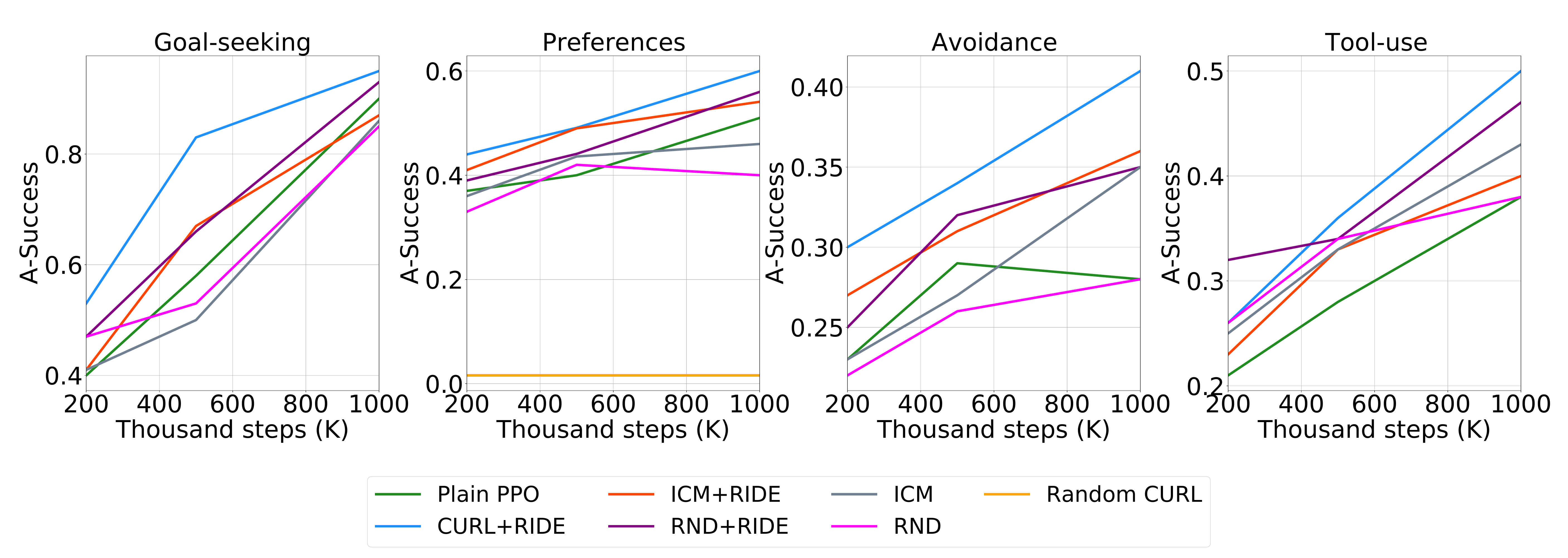}
\vspace{-2mm}
   \caption{A-success score of different models using 200K, 500K, and 1M interaction steps over the course of fine-tuning.}
 \vspace{-3mm}
\label{fig:efficiency}
\end{figure*}

\subsection{Experiment Setup}

\noindent \textbf{Implementation Details.} In all experiments, the agent only takes visual observations as input. The input of the network is an image of 84$\times$84 size. We set frame rate skip as  S$=$4 for all the tasks. To train our RL agent, we use PPO~\cite{schulman2017proximal} based on the PyTorch implementation.  We use a three-layer convolutional network to encode the image observations. During the open-ended exploration phase, we share the weights of the policy and model encoders. We find this strategy is important for a successful transfer to the downstream tasks.   During the evaluation phase, we use the weights of the pre-trained encoder as initialization and then fine-tuned the network to learn a task-specific policy with extrinsic rewards provided by the task.   For open-ended exploration, we set the initial learning rate as 1e-3, the environment switching threshold as 0.001, and the maximum interaction budget as 2 million.  For fine-tuning, we set learning rate as 2.5e-4. All evaluation tasks use 1 million interactions with extrinsic rewards for policy learning. 

\noindent \textbf{Reward function.}
We use different reward functions to train RL policies for each evaluation task. In the goal-seeking and tool-use tasks, we reward the agent with 1 if it hits the yellow ball, and otherwise 0. In the avoidance task, we only reward the agent with 1 if it hits the yellow sphere without entering the danger region. For the preference task, the agent receives a reward of 0.8 if it hits only the green ball, a reward of 0.2 if it hits only the yellow ball, and a reward of 1 if it hits both of them.

\noindent \textbf{Evaluation metrics.} 
We test all the agents on the same set of 100 randomly generated puzzles for each task. To quantify the performance of different agents, we adopt two metrics: mean return and A-Success score~\cite{bakhtin2019phyre} (i.e., mean returns weighted by number of actions taken).

\begin{itemize} %[leftmargin=*]
\item  \textit{Mean return} is defined as average returns over all 100 puzzles, given a task and budget of actions taken.  For tasks of goal-seeking, avoidance and tool-use, the return is either 0 or 1.  For preference task, the return for one episode could be 0, 0.2, 0.8 or 1. %We plot mean reward as function of attempt steps.

\item  \textit{A-Success}~\cite{bakhtin2019phyre} is a metric that jointly considers the mean returns and the number of actions taken. We use a weighted average of the returns under different action budgets to reflect how efficiently an agent solves a given task during the testing phase. Following~\cite{baker2019emergent}, we formulate it as 
\begin{equation}
    \begin{gathered}
        \texttt{A-Success} = \sum_{i}^{N} \frac{\alpha_{i} * s_{i}}{\sum_{j=1}^{N}\alpha_{j}} \ , \\
        \alpha_{i} = \log(i+1) - \log(i) \ , \ i\in \{ 1, 2, ..., N\}
    \end{gathered}
\end{equation}
where $s_{i}$ is the mean return (that is the average return across all the testing puzzles for each task) using $i$ actions, $\alpha_{i}$ is a weight for $s_{i}$, and $N$ denotes the maximum of action steps we consider. In this paper, we set $N=100$ for task 1, 2, 3, and $N=200$ for task 4. Since the weight $\alpha_i$ will decrease as the number of actions $i$ increases, solving a given task with fewer actions will lead to a higher A-success score. 
\end{itemize}

\subsection{Overall Performance}

% In our experiments, we mainly want to answer these following questions:

 %The varying level of difficulty of different tasks make the evaluation suite capable of assessing % systematically evaluating and comparing different agents' performance and tracking progress in this field.

For all the methods, we experiment with three different random seeds. Table~\ref{tab:result} summarizes the results measured by the A-success score for the different agents. From this table, we find that learning models without a task indeed can help downstream tasks, if we can design the algorithm appropriately.  For example,  CURL+RIDE outperform the baseline plain PPO for all the four downstream tasks.  %3) The performance gaps of the different models are larger for more challenging tasks. %These observations indicate that model-building is critical for tasks that require long-horizontal planning.  

We also plot the mean return curves for all seven agents on four downstream tasks in Figure~\ref{fig:overall} by averaging 3 runs. These curves show the mean return as a function of the number of actions performed for each given task. We calculate the return by averaging over all 100 testing puzzles in  OPEn. From these curves, we can see that the design of OPEn contains a diverse set of tasks of various difficulty levels.  For instance, it is relatively easy to solve the goal-seeking task within 100 interactions. However, in the challenging tool use task, which involves complex cognitive behaviors, i.e. using the ramp to go out of the fenced region, the mean return at 200 interactions is still not very high. 

\noindent \textbf{Human Evaluation.}
Two authors tested their performance on same set of 100 puzzles for each task. Their average score is summarized at the bottom of Table~\ref{tab:result}. We find that when using 1 million interactions for fine-tuning, the best model can achieve human-level performance on the basic goal-seeking task. However, for the other tasks, the performances of all RL models are significantly worse than the performance of the authors. These results indicate that current RL algorithms still struggle on physics reasoning tasks that require advanced cognitive skills.  

\subsection{Results Analysis}

\noindent \textbf{Are active explorations important for model-building?} In Table~\ref{tab:result}, we find that an agent (CURL+Random), which takes random actions to collect data, does not learn an effective representation for downstream tasks. The CURL+Random agent is even significantly worse when transferring the learned representation to the downstream tasks than the plain PPO model that is trained from scratch on the downstream tasks. These results provide validation for our claim that active exploration plays a vital role in learning models and representations that could be reused.

% \noindent \textbf{What is the most effective model for transfer learning?}
% As mention earlier, an agent that uses CURL for representation learning and RIDE for exploration achieved the best transfer learning results. Another interesting observations: RIDE rewards are more stable than state-prediction-based approaches. These observations are consistent with previous work~\cite{raileanu2020ride}.

\noindent \textbf{Sample efficiency for model-adaptation.}
We also examine the sample efficiency for model adaptation for different models. In figure~\ref{fig:efficiency}, we plot the curve of A-Success score as a function of the number of fine-tuning interactions with the environment. In particular, we evaluate different models trained with 200K, 500K, and 1 million interaction steps with the environment. We can find that most of the models do not learn meaningful skills using 200K step interactions. These results indicate that the representation learned with current model-building approaches are not highly sample efficient.% enough to directly transfer their model to the downstream tasks.

\section{Conclusions and Discussions}

We have introduced a new benchmark for open-ended learning in a physics environment. We examined several RL-based exploration methods and found a model learned through impact-driven exploration and unsupervised contrastive representation learning transfers better to the downstream tasks compared to alternatives. However, we also found that no agents could achieve human-level generalization and fast adaptation. We hope this benchmark can enable the development of model-building in open environments, and make progress toward agents that can learn general-purpose world knowledge that can be used for many tasks.

\vspace{5pt}
{\small \noindent\noindent \noindent \textbf{Acknowledgement:} This work is in part supported by ONR MURI N00014-16-1-2007, the Center for Brain, Minds, and Machines (CBMM, NSF STC award CCF-1231216), and IBM Research.}

% \section*{Broader Impact}
% Our work takes one step further toward a very big goal: general intelligence. We seek agents that build mental models of the world that are reusable across many tasks. These agents should not be trained with just a single task in mind since we want them to acquire general knowledge for many tasks. Developing a benchmark environment to evaluate the advantages and disadvantages of current algorithms systematically is an essential part of building more powerful and well-understood AI systems. Benchmarks like this can have the positive effect of clarifying progress and limitations in the field. At the same time, our evaluation suite is far from comprehensive, and may hide algorithmic deficiencies if not thoughtfully examined. For example, we have not attempted to measure robustness against adversarial perturbations in this benchmark, but that will be an important area for future extension, and especially for evaluating safety critical applications.

\bibliographystyle{plain}
\bibliography{egbib}

\end{document}

\maketitle
We first introduced the models and intrinsic rewards that we used for six baseline agents. We then provide details of the platform API. \textbf{Our environment and models will be open-source.}

\appendix

\section{Model Summary}

We summarize the models of representation learning and intrinsic rewards of explorations used in the 6 baseline agents. 

\begin{table}[h]
\centering
\begin{tabular}{lcccc}
Approach &  Model & Exploration \\
\toprule
ICM  & Forward + Inverse Dynamics & Forward + Inverse Dynamics\\
RND  &   State Prediction &  State Prediction \\
ICM + RIDE &  Forward + Inverse Dynamics & State Changes \\
RND + RIDE &  State Prediction & State Changes \\
CURL + RIDE &  Contrastive Prediction & State Changes \\
CURL + Random &  Contrastive Prediction & Random \\
    \bottomrule

\end{tabular}
\end{table}

\section{System API}

\subsection{Procedure generated sandbox environment}
\begin{verbatim}
def create_sandbox_environment():
    """
    Method for creating sandbox environment
    :return:
    """
    # We provide our environment through OpenAI Gym interface.
    # The environment can be run on remote server.
    # The env_ip is ip address of rendering server and
    # self_ip is server running this controller.
    # Port can be specified if Not specified default value is used
    env = gym.make('gym_open:open_puzzle_proc-v0',
                   env_ip="localhost",
                   self_ip="localhost",
                   port=None,
                   debug=False)

    # Example Puzzle configuration. This is used
    # for sandbox mode
    puzzle_config = {
        "main_sphere": 1,
        "main_sphere_i_j": [(8, 4)],
        "touch_sphere": 2,
        "cube": 2,
        "high_value_target": 1,
    }

    # To create a sandbox puzzle use 0, 1 and
    # puzzle configuration needs to be provided
    # To create one of 4 task scenes use 1, 1-4 instead
    env.add_change_puzzle(mode=0, task=0, puzzle_config=puzzle_config)
    env.set_observation(True)

    # Configure the environemnt by toggling
    # the action space and skip_frame
    env.configure_env(False, action_space="full", skip_frame=True)

    # Environment reward/ Extrinsic reward can
    # be toggle using this option
    env.turn_off_reward(True)
    done = False

    while not done:

        # Apply action and get observation, reward,
        # episode done and debug information.
        # The observation is dictionary containing
        # the image as well as scene object information
        obs, rewards, dones, debug = env.step(action)

        if done:

            # The enviormnet can be changed or simply
            # reset depending on requirement.
            env.add_change_puzzle(mode=0, task=0, puzzle_config=puzzle_config)
            obs = env.reset()
\end{verbatim}

\subsection{Procedure generated task environment}
\begin{verbatim}
def create_task_environment(task):
    """
    Method to create task specific environment
    :param task: Task number to create. Task has to be between 1 - 4
    :return: None
    """
    # We provide our environment through OpenAI Gym interface.
    # The environment can be run on remote server.
    # The env_ip is ip address of rendering server and
    # self_ip is server running this controller.
    # Port can be specified if Not specified default value is used
    env = gym.make('gym_open:open_puzzle_proc-v0',
                   env_ip="localhost",
                   self_ip="localhost",
                   port=None,
                   debug=False)

    # puzzle configuration needs to be provided
    # To create one of 4 task scenes use 1, 1-4 instead
    env.add_change_puzzle(mode=1, task=task)
    env.set_observation(True)

    # Configure the environemnt by toggling
    # the action space and skip_frame
    env.configure_env(False, action_space="full", skip_frame=True)

    done = False

    while not done:

        # Apply action and get observation, reward,
        # episode done and debug information.
        # The observation is dictionary containing
        # the image as well as scene object information
        obs, rewards, dones, debug = env.step(action)

        if done:

            # The enviormnet can be changed or simply
            # reset depending on requirement.
            env.add_change_puzzle(mode=1, task=task)
            obs = env.reset()
\end{verbatim}